\newif\iffinalversion
\finalversionfalse

\iffinalversion

\else

\fi

\documentclass[letterpaper, 10 pt, journal, twoside]{IEEEtran}
\usepackage{times}

\usepackage{graphics}           
\usepackage{times}              
\usepackage{amsmath}            
\usepackage{amssymb}            
\usepackage{graphicx}
\usepackage{algorithm}
\usepackage[noend]{algpseudocode}
\usepackage{booktabs}
\usepackage{color}
\definecolor{instructioncolor}{rgb}{.5,.5,.5}

\usepackage[font=small]{caption}

\def\secref#1{Sec.~\ref{#1}}
\def\figref#1{Fig.~\ref{#1}}
\def\tabref#1{Tab.~\ref{#1}}
\def\eqref#1{Eq.~(\ref{#1})}

\makeatletter
\usepackage{xspace}
\DeclareRobustCommand\onedot{\futurelet\@let@token\@onedot}
\def\@onedot{\ifx\@let@token.\else.\null\fi\xspace}
 
\def\ie{i.e\onedot}

\def\etal{{et al}\onedot}
\makeatother

\def\etalcite#1{\etal~\cite{#1}}

\usepackage{array}
\newcolumntype{L}[1]{>{\raggedright\let\newline\\\arraybackslash\hspace{0pt}}m{#1}}
\newcolumntype{C}[1]{>{\centering\let\newline\\\arraybackslash\hspace{0pt}}m{#1}}
\newcolumntype{R}[1]{>{\raggedleft\let\newline\\\arraybackslash\hspace{0pt}}m{#1}}

\newcommand{\RR}{\mathbb{R}}

\usepackage{siunitx}
\usepackage{booktabs}
\usepackage{graphicx}
\usepackage[numbers]{natbib}
\usepackage{multicol}
\usepackage[bookmarks=true]{hyperref}

\begin{document}

\title{Radar Instance Transformer: Reliable Moving Instance Segmentation in Sparse Radar Point Clouds}

\author{Matthias Zeller \IEEEmembership{Student Member, IEEE} \qquad \and Vardeep S. Sandhu\qquad \and Benedikt Mersch \IEEEmembership{Student Member, IEEE}\\ \and Jens Behley \IEEEmembership{Member, IEEE} \qquad\and Michael Heidingsfeld \qquad\and Cyrill Stachniss \IEEEmembership{Member, IEEE}%
  \thanks{Matthias Zeller and Vardeep S. Sandhu are with CARIAD SE and with the University of Bonn, Germany. Jens Behley and Benedikt Mersch are with the University of Bonn, Germany. Michael Heidingsfeld is with CARIAD SE, Germany. Cyrill Stachniss is with the University of Bonn, with the Department of Engineering Science at the University of Oxford, UK, and with the Lamarr Institute for Machine Learning and Artificial Intelligence, Germany.}%
}

\maketitle

\begin{abstract}
The perception of moving objects is crucial for autonomous robots performing collision avoidance in dynamic environments. LiDARs and cameras tremendously enhance scene interpretation but do not provide direct motion information and face limitations under adverse weather. Radar sensors overcome these limitations and provide Doppler velocities, delivering direct information on dynamic objects. In this paper, we address the problem of moving instance segmentation in radar point clouds to enhance scene interpretation for safety-critical tasks. Our Radar Instance Transformer enriches the current radar scan with temporal information without passing aggregated scans through a neural network. We propose a full-resolution backbone to prevent information loss in sparse point cloud processing. Our instance transformer head incorporates essential information to enhance segmentation but also enables reliable, class-agnostic instance assignments. In sum, our approach shows superior performance on the new moving instance segmentation benchmarks, including diverse environments, and provides model-agnostic modules to enhance scene interpretation. The benchmark is based on the RadarScenes dataset and will be made available upon acceptance.
\end{abstract}

\begin{IEEEkeywords} 
Radar Perception, Semantic Scene Understanding, Object Detection, Segmentation and Categorization, Deep Learning in Robotics and Automation
\end{IEEEkeywords}

\section{Introduction}
\label{sec:intro}

\IEEEPARstart{T}{he} safe navigation of autonomous vehicles in real-world environments depends on the reliable identification of moving objects. The knowledge about which part of the environment is moving and how many agents are present is essential for reliable future state prediction and path planning. This holds for mobile robots, offroad vehicles, and self-driving cars alike. Often, a redundant sensor setup of autonomous vehicles with different modalities is used and aims to reduce the risk of critical malfunctions. The advantages of individual sensors such as cameras, LiDARs, and radars should be considered to enhance the overall scene interpretation. Radar scans are affected by noise due to multi-path propagation, ego-motion, and sensor noise. At the same time, radar sensors are more robust to adverse weather conditions than cameras and LiDARs, thus essential for safe autonomous mobility. Furthermore, radar sensors measure the Doppler velocity and provide the radar cross section values, which depend on the material, the surface, and the geometry of the so-called radar detections, which could help to identify moving agents accurately~\cite{zeller2023ral}.

\begin{figure}[t]
  \centering
  \fontsize{9pt}{9pt}\selectfont
     \def\svgwidth{\linewidth}
     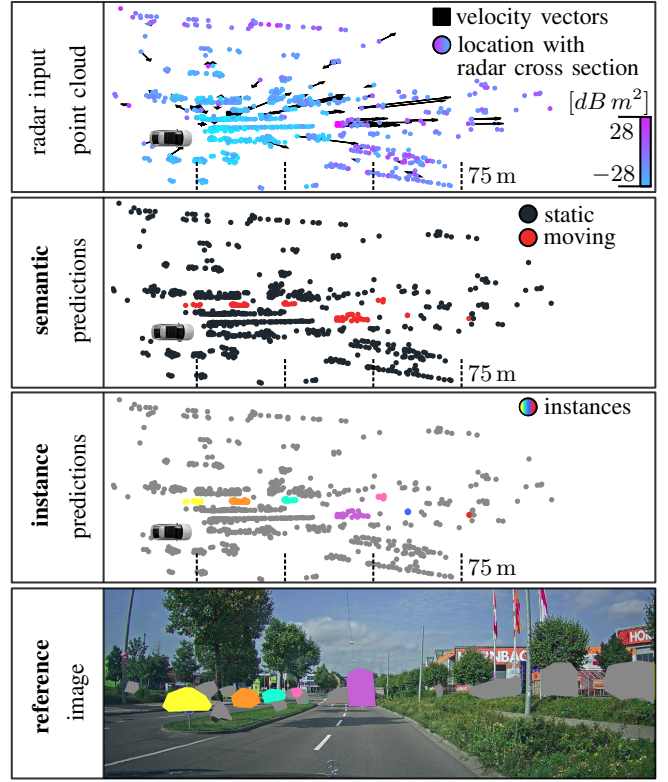
  \caption{Our method combines moving object segmentation (b) and instance segmentation (c) to solve the panoptic task of moving instance segmentation from sparse radar point clouds (a). The reference image in (d) illustrates the scene and includes privacy-preserving colored masks. In the image of the point cloud (c), each color represents a different instance of moving objects (static is grey). The colors in (c) and (d) correspond if the object is visible.}
  \label{fig:motivation}
  \vspace{-0.4cm}
\end{figure}
In this paper, we tackle the problem of moving instance segmentation in sparse and noisy radar point clouds. This requires differentiating between moving and static parts of the surrounding and separating individual agents. The task combines instance segmentation and moving object segmentation to a panoptic problem, including the differentiation of object instances, such as parked from moving cars. For many tasks, the information if an object is moving in combination with the velocities is often sufficient, substantially simplifying the labeling efforts. For radar data, the identification of the corresponding class is especially difficult and relies on synchronized LiDAR or camera data. However, semantic classes of the objects include additional information that can help to differentiate between nearby instances and enable class-dependent clustering. While semantic information could be exploited to reason about the mobility of objects, it also entails the so-called ``long-tail'' problem; \ie, no matter how much data we collect, we will always have classes that are underrepresented or not even covered by the training data. Moving or non-moving instances are essential for autonomous driving and allow it to potentially cover the aforementioned long-tail classes.

State-of-the-art methods often address the task of moving instance segmentation separately as moving object segmentation~\cite{chen2021ral} and instance segmentation~\cite{liu2022tiv}. Moving object segmentation approaches pass aggregated scans~\cite{chen2021ral,kim2022ral} through the whole network, which induces latency and is disadvantageous for a task requiring immediate feedback such as collision avoidance. Recent instance segmentation approaches~\cite{schult2022arxiv,vu2022cvpr,xie2021tro} work on single inputs, neglecting the temporal information, and do not differentiate between moving and static objects. Our approach aims to overcome this.

The main contribution of this paper is a novel approach that combines moving object segmentation and instance segmentation within a single network and accurately predicts moving instances in sparse and noisy radar point clouds. 
Our approach, called Radar Instance Transformer, predicts for each point in the input radar scan if it is moving or static and assigns an instance ID to each moving detection. To reliably identify moving instances, we efficiently incorporate the temporal information within the single current scan by our sequential attentive feature encoding module without passing aggregated scans through the whole network. We optimize the network architecture by processing the point cloud with the original resolution throughout the network to keep as much information as possible and enrich the points with high-level features. We utilize local and global attention to include instance information and propose a graph-based instance assignment to improve performance. In contrast to typical LiDAR approaches, our individual modules tackle the challenges of the sparsity and noise of radar point cloud properties that make radar data interpretation comparably hard. Besides, we extend the RadarScenes~\cite{schumann2021icif} dataset and transfer it into the first moving instance segmentation benchmark for point clouds.

In sum, we make five claims: First, our approach shows state-of-the-art performance for moving instance segmentation in sparse and noisy radar point clouds without passing multiple scans through the whole network. Second, our sequential attentive feature encoding extracts valuable temporal information by enriching the features of individual points to enhance accuracy. Third, our attentive instance transformer head is able to incorporate essential instance information, which improves the overall performance. Fourth, our attention-based graph partitioning enhances instance assignments without requiring class-dependent information. Fifth, our backbone enables valuable feature extraction for sparse radar point clouds by processing the full-resolution point cloud, \ie, the original number of points throughout the network.
\begin{figure*}[t]
 \centering
 \fontsize{8pt}{8pt}\selectfont
 \def\svgwidth{\textwidth}
 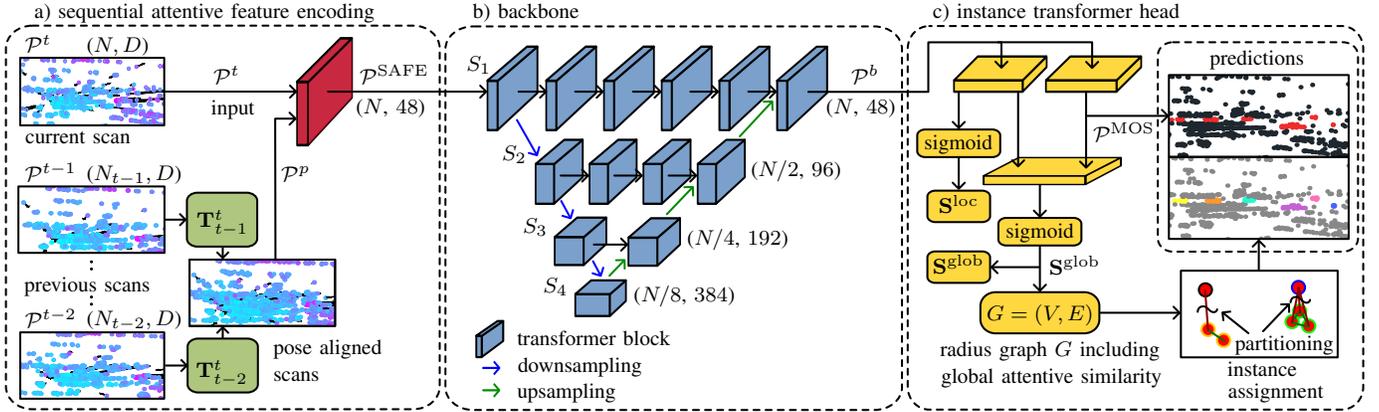%
 \caption{The detailed design of each module of our Radar Instance Transformer. (a) The sequential attentive feature encoding module combines the pose-aligned point clouds and their corresponding features using the current scan into a featurized point cloud, where the features encode temporal information. (b) This aggregated feature volume is then processed in a multi-pathway transformer backbone, where the transformer layers process the information at different resolutions and aggregate per-point features. (c) The per-point features are then used to estimate the semantics $\mathcal{P}^{\mathrm{MOS}}$ and perform graph partitioning for instance alignment based on the global similarity quantity $\mathbf{S}^{\mathrm{glob}}$. The tuples denote the number of points and feature channels. }
 \vspace{-0.1cm}
 \label{fig:modules}
  \vspace{-0.3cm}
\end{figure*}
\section{Related Work}
\label{sec:related}

Moving instance segmentation in point clouds unifies moving object segmentation~\cite{chi2022iros,kim2022ral,kreutz2023wacv,mersch2022ral,mohapatra2022visapp,qian2022neurips,sun2022iros} and instance segmentation~\cite{chen2021iccv,he2021cvpr,liang2021iccv,palffy2020ral,schult2022arxiv,vu2022cvpr} and can be solved as panoptic segmentation~\cite{li2022cvpr,razani2021iccv,sirohi2021tro}. Besides, the task also benefits from semantic segmentation~\cite{qian2022neurips,rosu2020rss,xin2022cvpr,xu2021iccv,xu2021cvpr,zhang2022ijis,zhao2021iccv} and scene flow estimation~\cite{baur2021iccvn,ding2022ral,ding2023cvpr,kittenplon2021cvpr,shi2022icra,wei2021cvpr}. To briefly summarize the extensive literature, including the different tasks, we distinguish between projection-based, voxel-based, point-based, and transformer-based approaches with and without considering temporal information.

\textbf{Projection-based} methods~\cite{chen2021ral,cortinhal2020svc,milioto2019iros,wu2018icra,wu2019icran} transfer the point cloud into frontal view images or 2D range images to exploit convolutional neural networks~(CNNs)~\cite{lecun1995ann}. These approaches enable effective processing of the 3D data and also leverage the tremendous architectural advances in image-based perception. Current approaches~\cite{chen2021ral,kim2022ral,sun2022iros} process the aggregated residual images of previous scans of 2D range images to include temporal information. The aggregation of scans results in repeating computations of individual scans throughout the whole network, which is computationally disadvantageous. Several approaches~\cite{mohapatra2022visapp,qiu2022tmlr,zhang2020cvpr,huang2022eccv} utilize a polar bird's-eye-view representation to enhance performance and account for the changing distribution of the points along the radial axis. However, projection-based methods face several limitations, including discretization artifacts and back projection errors, which can harm accuracy. The loss of information is impractical for radar data since detections are sparse.

\textbf{Voxel-based} methods~\cite{choy2019cvpr,qiu2022tmlr,zhu2021cvpr} mitigate back projection errors by maintaining 3D information of the point clouds. Since not all voxels are occupied, state-of-the-art methods such as Softgroup~\cite{vu2022cvpr}, SSTNet~\cite{liang2021iccv}, and HAIS~\cite{chen2021iccv} utilize submanifold sparse convolutional networks~\cite{graham2018cvpr} to extract discriminable features from spatially-sparse input data. Choy~\etalcite{choy2019cvpr} and Mersch~\etalcite{mersch2022ral} incorporate temporal information by aggregating scans over time and processing the point cloud throughout the network, which increases the computational burden. Recent methods~\cite{li2022cvpr,qiu2022tmlr} incorporate multiple representations, such as bird's-eye-view representation~\cite{zhang2020cvpr} with fine-grained voxel features to reduce runtime and improve accuracy. 
Despite the tremendous progress and encouraging results, voxelization inherently introduces discretization artifacts and information loss, which is especially harmful to the targeted processing of sparse radar point clouds~\cite{zeller2023ral}.

\textbf{Point-based} methods overcome the lossy encoding of aforementioned paradigms by directly processing the point clouds and keeping the spatial information intact. 
The pioneering work of Qi~\etalcite{qi2017cvpr} utilizes shared multi-layer perceptrons~(MLPs) to process the point cloud and aggregate nearby information to capture local structures by max pooling operations. Multiple approaches~\cite{puy2020eccv,schumann2018icif,wang2020wacv} adopt the hierarchical feature extraction from larger local regions of the successor, PointNet++~\cite{qi2017nips2}, to capture strong local structures in point clouds. Schumann~\etalcite{schumann2018icif} aggregate multiple radar point clouds for semantic segmentation and improve the approach by adding a memory module and additional features~\cite{schumann2020tiv}. In contrast, HARadNet~\cite{dubey2022mlwa} drastically restricts the region of interest to the near field to simplify the task and exploits additional properties to improve radar-based instance segmentation. 
However, for sparse outdoor data, the aggregation of multiple scans is key to including temporal information and extracting valuable features from the surrounding. Nevertheless, the processing of aggregated point clouds through the whole network comes with an increased computational burden. 

Another way to exploit stronger connections of the individual points is graph-based methods~\cite{landrieu2018cvpr,wang2019cvpr,zhao2019cvpr2} conducting message passing on the constructed graphs. Furthermore, graph convolutional networks~\cite{cheng2021cvpr,razani2021iccv} enable the extraction of valuable information within growing receptive fields. However, to reduce the computational burden, the graph representations use super-voxels~\cite{han2020cvpr2,shi2022cvpr} or over-segmented clusters~\cite{razani2021iccv} to combine local areas, which are inappropriate for sparse point clouds where single detection can represent complete instances.

In contrast, kernel-based convolutions~\cite{chi2022iros,fan2021cvpr,thomas2019iccv,xu2021cvpr} process point clouds on a per-point basis and extract features for individual points. State-of-the-art approaches~\cite{gasperini2021ral,kreuzberg2022eccvws} extend these methods with class-dependent post-processing steps to achieve competitive results, which are often inappropriate for moving instances segmentation, including different agents, without having different class labels.

\textbf{Transformer-based} methods exploit the self-attention mechanism~\cite{liu2022rss,vaswani2017nips,xie2018cvpr,yang2019cvpr} and achieve remarkable results within various tasks, from natural language processing~\cite{brown2020neurips,vaswani2017nips} to computer vision~\cite{carion2020eccv,cheng2022cvpr,kolesnikov2021iclr,yuan2021iccv} and point cloud understanding~\cite{guo2021pct,park2022cvpr,schult2022arxiv,shi2022icra,xin2022cvpr,zhang2022ijis,zhao2021iccv}.

Zhao~\etalcite{zhao2021iccv} utilize vector attention~\cite{zhao2020cvpr} to enable an individual weighting of the feature channels and propose trainable positional encoding~\cite{li2021nips}. Lai~\etalcite{xin2022cvpr} optimize the transformer architecture with a stratified key-sampling strategy to enlarge the effective receptive field, introduce contextual relative positional encoding to keep fine-grained position information throughout the network and use KPConv~\cite{thomas2019iccv} as the first layer embedding to enhance accuracy. Zeller~\etalcite{zeller2023ral} propose an optimized transformer architecture to extract features for semantic segmentation of sparse radar point clouds but only consider a single scan.

In contrast to the related work, we propose in this submission a novel architecture inspired by self-attention to efficiently include temporal information and conduct per-point attentive graph-based instance assignments to solve the new task of moving instance segmentation. 
Our Radar Instance Transformer captures complex structures in noisy and sparse point clouds and extracts valuable features by enriching the information of individual points of the current scan. Furthermore, we are able to include local and global instance information to achieve state-of-the-art performance in moving instance segmentation of radar data.

\section{Our Approach to Moving Instance Segmentation in Radar Data}
\label{sec:main}
Our goal is to achieve precise moving instance segmentation in a sequence of noisy, sparse radar point clouds to enhance scene interpretation of autonomous vehicles. \figref{fig:modules} illustrates the overall architecture of our Radar Instance Transformer, which is a point-based framework and builds upon the self-attention mechanism~\cite{vaswani2017nips}. We incorporate temporal relationships in a computationally efficient way to enrich the features of the current point cloud, as detailed in~\secref{sec:safe}.
We furthermore propose a new backbone, processing the full-resolution point clouds throughout the network to extract valuable features and omit information loss in~\secref{sec:backbone}. We utilize local and global information derived from self-attention to optimize the panoptic task of instance association and semantic segmentation, which we explain in~\secref{sec:mith}. Finally, our graph-based instance assignment, detailed in~\secref{sec:gia}, incorporates global instance information to enhance accuracy.

\subsection{Sequential Attentive Feature Encoding}
\label{sec:safe}
The temporal information is essential to enhance scene interpretation in sparse and noisy radar point clouds. Especially for moving instance segmentation, the temporal relationship helps to identify agents reliably. Additionally, temporal dependencies support the differentiation of moving and static detections, including noise directly definable in the temporal domain due to the often changing appearance in sequences of scans. In contrast to other approaches~\cite{kim2022ral,mersch2022ral} that pass multiple point clouds through the whole network, we propose the sequential attentive feature encoding~(SAFE) module to efficiently enrich the features of a single point cloud with temporal information. Therefore, we only pass the current scan through the whole network and do not increase the number of points that need to be processed by the network, whereas the previous scans are only processed within the SAFE module. The goal is to adaptively combine the temporal information of the previous scans with the features of the current point cloud, hence reducing the computational burden but still keeping important information.

The input to our SAFE module is the current scan $\mathcal{P}^{t}$ at time~$t$ and $T$ previous scans $\mathcal{P}^{t-T},\dots,\mathcal{P}^{t-1}$ of a sequence of radar scans, as depicted in~\figref{fig:modules}. The current scan $\mathcal{P}^{t}$ includes the point coordinates $\mathbf{P}^t=[\mathbf{p}_1, \dots, \mathbf{p}_N]^{\top} \in \RR^{N\times 3}$ and point-wise features $\mathbf{X}^t=[\mathbf{x}_1,\dots,\mathbf{x}_{N}]^{\top} \in \RR^{N\times D}$, where $\mathbf{p}_i\in\RR^{3}$ and $\mathbf{x}_i\in\RR^{D}$ for $N$ points. The point-wise features comprise the radar measurements for the individual detections, including velocity, position, and radar cross section, resulting in the input feature dimension $D$. We aggregate the previous scans into a single scan, which comprises the ${N}_{t-T}+\cdots+N_{t-1}$ points of all the combined $T$ scans with point coordinates $\mathbf{P}^{p}$ and point-wise features $\mathbf{X}^p$. We keep the original number of points for each individual point cloud. To align the previous scans with the current scan locally, we assume that the homogeneous relative pose transformation $\mathbf{H}^{t}_{t-1}\in \RR^{4\times 4}$ between point cloud $\mathcal{P}^{t}$ and $\mathcal{P}^{t-1}$ is given. The alignment helps to combine the information of the related instances over time and supports the identification of noise within a sequence of scans. The transformation can be derived from GPS, online SLAM algorithms~\cite{adolfsson2023tro,behley2018rss,campos2021tro,zhang2014rss}, or using wheel encoders and inertial measurement units. In our work, we use the provided vehicle odometry and DGPS information.

Since the data association of individual radar points within unprocessed point clouds is difficult~\cite{xiao2021neurips,xin2022cvpr}, we first process the point-wise features of the current scan $\mathbf{X}^t$ and the features of the previous scans $\mathbf{X}^p$ with a KPConv layer~\cite{thomas2019iccv} to extract higher dimensional features $\mathbf{x}_i \in\RR^{D_1}$. The aggregated features of the previous scans and the features of the current scan are processed by a separate KPConv layer. The resulting features are still point-wise features with dimension ${D_1}$. To adaptively combine the features and extract temporal information, we introduce an intra-attention module that is inspired by the Point Transformer layer~\cite{zhao2021iccv}. We first encode the features of the previous scans $\mathbf{X}^{p}$ as values $\mathbf{V}$ and keys $\mathbf{K}$ and the features of the current scan $\mathbf{X}^{t}$ as queries $\mathbf{Q}$ as follows:
\begin{align}
\mathbf{Q} &= \mathbf{X}^t \mathbf{W}_Q\text{,} &
\mathbf{K} &= \mathbf{X}^p \mathbf{W}_K\text{,} &
\mathbf{V} &= \mathbf{X}^p \mathbf{W}_V\text{.}
\label{eq:1}
\end{align}

The matrices $\mathbf{W}_{Q} \in \RR^{D_1\times D_2}$, $\mathbf{W}_{K} \in \RR^{D_1\times D_2}$, and $\mathbf{W}_{V} \in \RR^{D_1\times D_2}$ are learned linear projections. We increase the feature dimension of the encoding to $D_2$ to enable fine-grained information aggregation. Since temporal information is present within the local neighborhood, we restrict the intra-attention of the current scan and the aligned previous scans to local areas. To determine the local areas, we calculate the $k$-nearest neighbors~($k$NN) with $k=N^l$ for the points in the current point cloud $\mathcal{P}^{t}$ within the aligned past point clouds $\mathcal{P}^{p}$. This results in the combination of the respective detections over time. To extract the related information of the queries, values, and keys within the local areas, we utilize the sample and grouping algorithm~\cite{qi2017nips2} resulting in $\mathbf{Q}^{*},\mathbf{K}^{*}$, and $\mathbf{V}^{*} \in \RR^{N \times N^l \times D_2}$. The queries comprise the repeated entries of the current point cloud, and the keys and values include the encoded features of the past point clouds, which belong to the corresponding local neighbors of the $N$ points of the current point cloud. To further include the relative positional encoding~\cite{zhao2021iccv}, we determine the relative positions within the local areas of the two point clouds where $\mathbf{p}_i \in \mathcal{P}^{t}$ and $\mathbf{p}_j \in \mathcal{P}^{p}$ resulting in:
\begin{align}
\mathbf{r}_{i,j}=\mathbf{p}_i-\mathbf{p}_j.
\label{eq:r}
\end{align}

Based on the relative positions $\mathbf{r}_{i,j}$, we compute the positional encoding $\mathbf{R} \in \RR^{N \times N^l \times D_2}$ by an MLP, including two linear layers with weight matrices $\mathbf{W}_{R_1} \in \RR^{3\times 3}$ and $\mathbf{W}_{R_2} \in \RR^{3\times D_2}$, batch normalization~\cite{ioffe2015arxiv}, and rectified linear unit~(ReLU) activation function~\cite{nair2010icml}.
To calculate the attention weights $\mathbf{A} \in \RR^{N \times N^l \times D_2}$ for the individual points $i$, we adopt vector attention~\cite{zhao2020cvpr} and subtract the encoded keys of the past point clouds from the encoded queries of the current point cloud where $\mathbf{Q}^{*}_{i},\mathbf{K}^{*}_{i} \in \RR^{N^l \times D_2}$. Similarly, we add the relative positional encoding $\mathbf{R}_{i}$ to include fine-grained position information in the weighting. Hence, the attention weights connect the detections in the current scan with the detections in the aligned previous scans, which we determine as follows:
\begin{align}
\mathbf{A}_{i}= \text{softmax}((\mathbf{Q}^{*}_{i}-\mathbf{K}^{*}_{i})+\mathbf{R}_{i}),
\label{eq:4}
\end{align}
where we apply the softmax function to each feature of the points within the local area individually. We process the resulting attention weights by a multi-layer perceptron with two fully connected layers, where each layer is followed by batch normalization~\cite{ioffe2015arxiv}.
To aggregate the weighted features $\mathbf{X}^{\mathrm{temp}}$, which comprise the temporal information of previous scans encoded as values, we calculate the sum of the element-wise multiplication, indicated by $\odot$, as:
\begin{align}
\mathbf{X}^{\mathrm{temp}}_{i} &= \sum_{j=1}^{N^{l}}{ \mathbf{A}_{i,j}\odot (\mathbf{V}^{*}_{i,j}+\mathbf{R}_{i,j}) }.
\label{eq:5}
\end{align}

Additionally, we include the positional encoding~\cite{zhao2021iccv} because it provides valuable information about the local areas and their appearance over time. Static objects remain in the same location, whereas moving objects change position within a sequence of scans, which both can be represented within the positional encoding. 
To combine the information within the current scan, we concatenate the temporal features $\mathbf{X}^{\mathrm{temp}}\in \RR^{N \times D_2}$ and the features of the current scan $\mathbf{X}^{t} \in \RR^{N \times D_1}$ after the KPConv layer resulting in the temporal enriched features $\mathbf{X}^{\mathrm{SAFE}}\in \RR^{N \times (D_1 + D_2)}$ of our SAFE module. Besides the features, the output point cloud $\mathcal{P}^{\mathrm{SAFE}}$ includes the point coordinates $\mathbf{P}^{\mathrm{SAFE}}$, where $\mathbf{P}^{\mathrm{SAFE}}=\mathbf{P}^{t}$. The point coordinates are not processed to include fine-grained position information within the consecutive network. The SAFE module is model agnostic and can be applied to different backbones.
\subsection{Backbone}
\label{sec:backbone}
The motivation behind our backbone design is to keep the full resolution, \ie, the original number of points throughout the network, and thus avoid information loss and extract valuable features from sparse radar scans, as illustrated in~\figref{fig:modules}. Contrary to the widely-used U-Net~\cite{qi2017nips2,zhao2021iccv} with an encoder-decoder architecture including skip connection, we process the full-resolution point cloud in parallel to the downsampled feature maps and enrich the information in the original point cloud with high-level features. In contrast to 3D LiDAR scans, the sparsity of radar data enables processing full-resolution point clouds but also requires keeping as much information as possible to enhance performance. To handle the changing number of points in radar scans, we adopt the advanced batching algorithm~\cite{zhao2021iccv} to always keep the original number of points. 
Compared to the first parallel network for image analysis, the HRNet~\cite{wang2020pami}, we reduce the processing of the high-level features to a minimum and do not keep the parallel tracks throughout the network. Considering that the processing of these features is computationally expensive, HRNet reduces the input size to keep the parallel feature extraction, which is in contrast to our approach that preserves the information of the point cloud.

The input point cloud $\mathcal{P}^{\mathrm{SAFE}}$ of our backbone includes the temporally enriched features $\mathbf{X}^{\mathrm{SAFE}}$ and the position information~$\mathbf{P}^{t}$ of the current scan. The dimension of the features for the stages $S_L$ with $L= 1,\dots, 4$ are $D_{S_L}$, where $D_{S_1}=(D_1 + D_2)$, $D_{S_2}=2D_{S_1}$, $D_{S_3}=2D_{S_2}$, and $D_{S_4}=2D_{S_3}$. The individual building blocks of our backbone are the commonly used transformer blocks as well as up- and downsampling layers. The stages $S_1$, $S_2$, $S_3$, and $S_4$ comprise 6, 4, 2, and 1 transformer blocks, respectively. The transformer block is a residual block that follows the design of the Stratified Transformer block~\cite{xin2022cvpr}, where we replace the window attention with the Point Transformer layer~\cite{zhao2021iccv} since it is computationally more efficient. Within the block, we first process the input features by layer normalization~\cite{ba2016arxiv} before feeding the features and position information into the transformer layer. The resulting output is processed by two fully connected layers, including the Gaussian error linear unit~\cite{hendrycks2016corr} activation function and layer normalization. We add the features of the residual path to the updated features to determine the output features. 

The transformer layer adopts the architecture of the Point Transformer layer~\cite{zhao2021iccv} and follows the implementation of our SAFE module. We utilize the inter-attention operation and encode the features $\mathbf{X}^{\mathrm{SAFE}}$ as values, keys, and queries following~\eqref{eq:1}. In contrast to our SAFE module, we do not increase the dimension of the encodings by the learned linear projections, hence we set the dimension $D_1$ and $D_2$ of $\mathbf{W}_{Q}$, $\mathbf{W}_{K}$, and $\mathbf{W}_{V}$ to $D_{S_L}$. We determine the local areas within the current point cloud based on the $k$NN algorithm with $k=N^l$. To align the queries and keys within the local areas, we utilize the sample and grouping algorithm, as mentioned in~\secref{sec:safe}. We calculate the positional encoding by an MLP, including two linear layers, batch normalization~\cite{ioffe2015arxiv}, and ReLU activation function~\cite{nair2010icml}. We adjust the output dimension of $\mathbf{W}_{R_2}$ to $D_{S_L}$. The relative position changes to $\mathbf{r}_{i,j}=\mathbf{p}_i-\mathbf{p}_j$ where $\mathbf{p}_j, \mathbf{p}_i \in \mathcal{P}^{t}$, to include detailed position information of the current scan. We calculate the final output features of the transformer layer based on~\eqref{eq:5}. The point coordinates are not further processed to keep the information intact and improve the features with fine-grained position information within consecutive blocks. 

For the downsampling, we utilize the commonly used farthest point sampling with $k$ nearest neighbor~($k$NN) search with $k=N^d$ and max pooling~\cite{qi2017nips2}. The top level does not include downsampling since we want to keep the original number of points. From stage $S_L$ to stage $S_{L+1}$, we sample $N_{S_{L+1}}$ points from $N_{S_L}$ points where $N_{S_{L+1}}=N_{S_L}/2$. We reduce the cardinality of the point cloud by a factor of 2 instead of 4~\cite{qi2017nips2,zhao2021iccv} to keep more information since radar point clouds are sparse, resulting in $[N, N/2, N/4, N/8]$ points for the four stages, where $N$ is the number of points in the current scan. Based on the local areas, we aggregate the features and process them by a linear layer with weight matrix $\mathbf{W}^{\mathrm{down}}_{S_L} \in \RR^{D_{S_L} \times D_{S_{L+1}}}$ where $D_{S_{L+1}}=2 \cdot D_{S_L}$ and layer normalization. Afterward, we apply max pooling to derive the high-level features of the downsampled point cloud and pass the features and the point coordinates of the sampled points to the consecutive stage. 

For the upsampling, we adopt the trilinear interpolation based on an inverse distance weighted average~\cite{qi2017nips2}. We determine $k=3$ neighbors of the points of stage $S_{L-1}$ within the downsampled points of stage $S_{L}$ based on the $k$NN algorithm. We process the features of stage $S_L$ by a linear layer with $\mathbf{W}^{\mathrm{up}}_{S_L} \in \RR^{D_{S_L} \times D_{S_{L-1}}}$ where $D_{S_{L-1}}=D_{S_L}/ 2$ and layer normalization to adjust the dimensions of the features of the two stages. We interpolate the resulting features and add them to the features of the current stage $S_{L-1}$, which we additionally process by a linear layer with $\mathbf{W}^{\mathrm{up}}_{S_{L-1}} \in \RR^{D_{S_{L-1}} \times D_{S_{L-1}}}$ and layer normalization. The upsampled features are processed within the succeeding stages to integrate the information iteratively. Subsequently, we add high-level features to the full-resolution point cloud of stage $S_1$ to derive the final output features of our backbone $\mathbf{X}^b \in \RR^{N \times D_{S_1}}$. The advanced network architecture enables the extraction of valuable features for sparse radar point clouds to enhance performance.
\subsection{Moving Instance Transformer Head}
\label{sec:mith}
Our moving instance transformer head combines moving object segmentation with instance segmentation to derive the final panoptic output, as detailed in~\figref{fig:modules}. For moving instance segmentation, currently moving agents, such as cars, bikes, and trucks, belong to the same moving class without further differentiation. For many tasks, the information of being a moving object in combination with the velocities is often sufficient, simplifying the labeling efforts. However, state-of-the-art approaches for panoptic segmentation and instance segmentation rely on specific semantic labels to enhance accuracy~\cite{chen2021iccv,gasperini2021ral,li2022cvpr,vu2022cvpr}. To overcome current limitations and further incorporate instance information, we propose a new attention-based module to reliably identify moving instances in sparse and noisy radar point clouds.

First, we predict the per-point moving object segmentation~(MOS) class $\mathcal{P}^{\mathrm{MOS}}=\{p^{\mathrm{MOS}}_1,\dots,p^{\mathrm{MOS}}_N\}$, where $p^{\mathrm{MOS}}_i\in \{\mathrm{static},\mathrm{moving}\}$ using a multi-layer perceptron. To identify points belonging to the same instances, our idea is to deduce an attentive similarity quantity $\mathbf{S}_{i,j} \in \RR$ based on the self-attention mechanism. Therefore, we process the output features of the backbone $\mathbf{X}^b$ by two linear layers to encode the features as keys $\mathbf{K}^b$ and queries $\mathbf{Q}^b$, with $\mathbf{W}^b_{Q} \in \RR^{D_{S_1}\times D_{S_1}}$ and $\mathbf{W}^b_{K} \in \RR^{D_{S_1}\times D_{S_1}}$ following~\eqref{eq:1}. Since global attention is expensive to compute within the complete point cloud, we restrict the computation to local areas, which we determine by $k$NN with $k=N^a$. We aggregate the local neighborhoods by sample and grouping, resulting in $\mathbf{Q}^{b*},\mathbf{K}^{b*} \in \RR^{N \times N^a \times S_1}$, as explained in~\secref{sec:safe}. We calculate the relative positions within the resulting local areas following~\eqref{eq:r}, where $\mathbf{p}_i$, $\mathbf{p}_j \in \mathcal{P}^{t}$. We reduce the dimensionality of the relative positional encoding $\mathbf{R}^b$ by a linear layer with weight matrix $\mathbf{W}^b_{R} \in \RR^{3 \times 1}$ and ReLU activation function~\cite{nair2010icml} to include the position information. To derive a scalar value for the local similarity quantity, we calculate the dot-product attention $\mathbf{A}^{\mathrm{loc}}\in \RR^{N \times N^a}$ for the keys and queries, resulting in:
\begin{align}
\mathbf{A}^{\mathrm{loc}}_{i,j}=&\mathbf{Q}^{b*}_{i,j}\mathbf{K}^{b*}_{i,j}{}^{\top}+\mathbf{R}^b_{i,j}.
\label{eq:dot}
\end{align}

The goal is to have local similarity quantities close to 1 for points within the local area that belong to the same instance and values close to 0 for different instances. Therefore, we replace the softmax function with the sigmoid function to determine the final local similarity quantity $\mathbf{S}^{\mathrm{loc}}$ as:
\begin{align}
\mathbf{S}^{\mathrm{loc}}_{i,j}=\mathrm{sigmoid}(\mathbf{A}^{\mathrm{loc}}_{i,j})=&\frac{1}{1+\exp(-\mathbf{A}^{\mathrm{loc}}_{i,j})}.
\label{eq:sig}
\end{align}

The resulting local similarity matrix $\mathbf{S}^{\mathrm{loc}} \in \RR^{N\times N^a}$ includes a similarity quantity for all $N$ points of the current point cloud. The ground truth of the local similarity matrix can be directly derived from the indices of the $k$NN algorithm and the ground truth instance IDs. Since the focus is to extract valuable information for moving instances, we set the similarity quantity for points within the local area that both belong to the static class to 0, which we elaborate in more detail in~\secref{sec:abltranshead}.

The restriction to local areas induces the problem that often not all points that belong to one instance have an attentive similarity quantity. To overcome this limitation, we introduce the global attentive similarity quantity $\mathbf{S}^{\mathrm{glob}}$ for moving detections. Hence, the second part of our instance transformer head combines the prediction of the moving object segmentation and the approach of our local similarity quantity. Since computing the global similarity, including all points, comes with a large computational burden, we first select the $N^{\mathrm{mov}}$ points that we predict as moving. In contrast to Cheng~\etalcite{cheng2022cvpr}, we select all moving predictions, not just those predicted within the same mask. We already encoded the features as keys $\mathbf{K}^b$ and queries $\mathbf{Q}^b$ for the local attentive similarity such that no additional encoding is required. We sample and group the global moving predictions for the queries and keys, resulting in $\mathbf{Q}^{b**},\mathbf{K}^{b**} \in \RR^{N^{\mathrm{mov}} \times N^{\mathrm{mov}} \times S_1}$. We use the dot-product to derive the global attention weights $\mathbf{A}^{\mathrm{glob}}\in \RR^{N^{\mathrm{mov}} \times N^{\mathrm{mov}}}$ as:
\begin{align}
\mathbf{A}^{\mathrm{glob}}_{i,j}=&\mathbf{Q}^{b**}_{i,j}\mathbf{K}^{b**}_{i,j}{}^{\top}.
\label{eq:dot2}
\end{align}

We remove the relative positional encoding because the values largely differ in magnitude and, at least in our experience detailed in~\secref{sec:abltranshead}, do not provide important information to differentiate instances in the global context. We follow~\eqref{eq:sig} to calculate the final global similarity quantity $\mathbf{S}^{\mathrm{glob}} \in \RR^{N^{\mathrm{mov}}\times N^{\mathrm{mov}}}$. We construct the ground truth matrix for the global similarity matrix based on the semantic predictions of the moving instance detections. 
The attentive similarity modules are differentiable and can be learned in an end-to-end manner. We emphasize that our attention-based similarity quantity encodes essential information for moving instance association and moving object segmentation in noisy radar point clouds. Further, the local attention mechanism includes the noise in the ground truth and directly penalizes the association of noise with an instance, which is beneficial for radar data processing. Incorporating global information supports the extraction of discriminative features, especially for sparse point cloud processing. The computation of the local and global similarity quantities is model agnostic and can be combined with different backbones. 
\subsection{Graph-based Instance Assignment}
\label{sec:gia}
The final part of our instance transformer head is the graph-based instance assignment module. It utilizes the predictions $\mathcal{P}^{\mathrm{MOS}}$ and the global similarity matrix $\mathbf{S}^{\mathrm{glob}}$, which includes a similarity quantity for each pair of moving predictions to derive the final instance IDs. The key idea is to overcome the limitations of the direct assignment since the global similarity matrix may include multiple instance assignments and different similarity quantities for different instances, which are difficult to solve. Furthermore, we want to remove the class dependencies mentioned in~\secref{sec:mith} to derive the final instances. 

We first construct a radius graph $G = (V,E)$ based on the prediction of moving detections and the position information to take into account that sparse and noisy radar point clouds differ in density. The moving points represent the vertices $V=\{1,\dots, N^{\mathrm{mov}}\}$ of the graph. We derive the set of $m$~edges~$E$ by connecting each moving point with the neighboring points, which lie within the spatial radius~$r$. The resulting graph can be represented by the square adjacency matrix $\mathbf{A}^{\mathrm{adj}} \in \RR^{N^{\mathrm{mov}}\times N^{\mathrm{mov}}}$, where $\mathbf{A}^{\mathrm{adj}}_{i,j} = 1$, if $||p_i-p_j|| \le r $ and otherwise $\mathbf{A}^{\mathrm{adj}}_{i,j} = 0$. To derive the instance IDs, we partition the graph into $V_1,\dots, V_{k'}$ disjoint subgraphs, where $k'$ is not given in advance because the number of instances within a scan is unknown.
Based on the constructed graph, our idea is that individual instances have highly interconnected nodes and points belonging to different instances are sparsely connected. To exploit this property, we partition the graph by maximizing the modularity via a spectral approach~\cite{newman2006pnas}. The modularity depends on the number of edges falling within clusters minus the approximated expected number of random edges in an equivalent network. For the case of two clusters and $s_i = 1$, if vertex $i$ belongs to the first cluster and $s_i = -1$ if it belongs to the second one. The modularity is: 
\begin{equation}
Q = \frac{1}{4m} \sum_{i,j} \biggl( \mathbf{A}^{\mathrm{adj}}_{i,j} - \frac{k_ik_j}{2m} \biggr) s_i s_j\text{,}
\label{qvalue}
\end{equation}
where $k_i$ and $k_j$ are the degrees, the number of edges incident to a vertex, of the corresponding vertices. The adjacency matrix $\mathbf{A}^{\mathrm{adj}}_{i,j}$ contains information about the actual number of edges, and the factor $\frac{k_ik_j}{2m}$ is the approximated expected number of edges between vertices $i$ and $j$ if the edges are placed randomly. The modularity can be further expressed by the modularity matrix \mbox{$B_{i,j} = \mathbf{A}^{\mathrm{adj}}_{i,j} - \frac{k_ik_j}{2m}$} as:
\begin{equation}
Q =\frac{1}{4m}{\mathbf{s}}^{\top}\mathbf{B}{\mathbf{s}}\text{,}
\label{eq:qq}
\end{equation}
where $\mathbf{s}$ is the vector of the elements $s_i$.
To correctly divide the network into more than two clusters, we construct a modularity matrix $\mathbf{B}^{\mathrm{sub}}$ for each subgraph resulting in:
\begin{equation}
\mathbf{B}^{\mathrm{sub}}_{i,j} = \mathbf{A}^{\mathrm{adj}}_{i,j} - \frac{k_ik_j}{2m}
               - \delta_{i,j} \biggl[ k_i^{\mathrm{sub}} - k_i \frac{d^{\mathrm{sub}}}{2m} \biggr]\text{,}
\label{defsbg}
\end{equation}
where $k_i^{\mathrm{sub}}$ is the degree of the node $i$ within the subgraph, $d^{\mathrm{sub}}$ is the sum of all degrees $k_i$ of the nodes in the specific subgraph, and $\delta_{i,j}$ is the Kronecker delta, which is 1 if both nodes belong to the same subgraph and zero otherwise. We calculate the subgraph modularity $Q^{\mathrm{sub}}$ following~\eqref{eq:qq} to determine the additional contribution to the total modularity~$Q$. 

The modularity matrix $\mathbf{B}$ always depends on the adjacency matrix $\mathbf{A}^{\mathrm{adj}}$. Since the plain radius graph leads to interconnections of nearby instances, represented in $\mathbf{A}^{\mathrm{adj}}$, we optimize the cluster assignment by including additional information about the vertices. Our global attentive similarity matrix $\mathbf{S}^{\mathrm{glob}}$ directly provides this valuable information since the similarity quantities differentiate between instances. Hence, we calculate the element-wise product of both matrices as: 
\begin{equation}
\mathbf{A}^{\mathrm{adj}}_{\mathrm{attn}}= \mathbf{S}^{\mathrm{glob}} \odot \mathbf{A}^{\mathrm{adj}}\text{,} 
\label{eq:qq2}
\end{equation}
to resolve the issue and derive the final attention-based adjacency matrix $\mathbf{A}^{\mathrm{adj}}_{\mathrm{attn}}$ for the instance assignment. The modularity, which now incorporates the additional instance information, is maximized by the vertex moving method to reach the best possible modularity. The optimization stops if any proposed split for all the subgraphs has a negative or no effect on the modularity. The final partitions represent our instance IDs. The partitioning is class independent and incorporates attention-based instance information.

\section{Implementation Details}
\label{sec:impl}
We implemented our Radar Instance Transformer in PyTorch~\cite{paszke2019nips} and train our network with one Nvidia A100 GPU and a batch size $b$ of 64 over 100 epochs. We utilize the AdamW~\cite{loshchilov2017iclr} optimizer with an initial learning rate of 0.001 and drop the learning rate by a factor of 10 after 60 and 80 epochs to train our model. We use the binary cross-entropy loss to learn the local $\mathbf{S}^{\mathrm{loc}}$ and global $\mathbf{S}^{\mathrm{glob}}$ similarity matrix for instance association. For the semantic output, we utilize the focal Tversky loss~\cite{abraham2019isbi}.

The individual scans, which are the input to our network, are sparse radar point clouds with $N$ points and feature dimension $D$. The RadarScenes~\cite{schumann2021icif} dataset comprises two spatial coordinates and the Doppler velocity, whereas the View-of-Delft dataset~\cite{palffy2022ral} also includes the elevation information. The radar sensors are often defined as 2+1D and 3+1D radars, respectively. Each sequence consists of $T+1$ scans, and the batch size $b$ is the number of input sequences. For RadarScenes, we extend the coordinates $x^C_{i}$, $y^C_{i}$ with the coordinate $z^C_{i}=0$ to apply the pose transformation mentioned in~\secref{sec:safe}. In addition to the position information and the Doppler velocity $v_i$, the radar sensors provide the radar cross section $\sigma_i$, resulting in input vector $\mathbf{x}_i=(x^C_i, y^C_i, z^C_i, \sigma_i, v_i)$.

We first increase the per-point features to $D_1=16$ before we apply the intra-attention. We increase the dimension in our SAFE module and set $D_2=32$ resulting in an output feature dimension of 48, which is kept for the full-resolution stage $D_{S_1}=48$ and gradually increased to $D_{S_2}=96$, $D_{S_3}=192$, and $D_{S_4}=384$ during downsampling. We set $N^l=N^d=N^a=12$, where $N^l$ represents the local area in the transformer layers, $N^d$ the local neighborhood for downsampling, and $N^a$ the local area for the attentive local similarity quantity. The input sequence consists of the current point cloud and $T=2$ previous scans, which we further elaborate in~\secref{sec:absafe}. We pad missing previous scans as zeros to always include $T+1$ scans as input. The padded scans comprise a set of $1,024$ points where we set the features to zero. The padding is adopted until $T+1$ scans are available. To reduce overfitting during training, we apply data augmentation~\cite{xu2021cvpr}, including Gaussian jittering (variance of 0.01), scaling [$\times$0.95, $\times$1.05], shifting ($\pm$\SI{0.1}{m}), and random flipping of the point cloud. 
\section{Experimental Evaluation}
\label{sec:exp}
The main focus of this work is to achieve reliable moving instance segmentation in sparse and noisy radar point clouds.
We present our experiments to show the capabilities of our method and to support our key claims, including that our approach achieves state-of-the-art performance in moving instance segmentation without processing multiple scans throughout the whole network. Our sequential attentive feature encoding module improves accuracy by efficiently incorporating temporal information. In addition, the attentive instance transformer head enhances the overall performance of moving instance segmentation by including local and global instance knowledge. Our attention-based graph clustering enhances class-independent instance assignments. The optimized backbone architecture extracts discriminable features from sparse radar point clouds by enriching individual detection with high-level features.
\begin{figure*}[t]
 \centering
 \fontsize{8pt}{8pt}\selectfont
 \def\svgwidth{0.95\textwidth}
 \begingroup%
  \makeatletter%
  \providecommand\color[2][]{%
    \errmessage{(Inkscape) Color is used for the text in Inkscape, but the package 'color.sty' is not loaded}%
    \renewcommand\color[2][]{}%
  }%
  \providecommand\transparent[1]{%
    \errmessage{(Inkscape) Transparency is used (non-zero) for the text in Inkscape, but the package 'transparent.sty' is not loaded}%
    \renewcommand\transparent[1]{}%
  }%
  \providecommand\rotatebox[2]{#2}%
  \newcommand*\fsize{\dimexpr\f@size pt\relax}%
  \newcommand*\lineheight[1]{\fontsize{\fsize}{#1\fsize}\selectfont}%
  \ifx\svgwidth\undefined%
    \setlength{\unitlength}{505.89001465bp}%
    \ifx\svgscale\undefined%
      \relax%
    \else%
      \setlength{\unitlength}{\unitlength * \real{\svgscale}}%
    \fi%
  \else%
    \setlength{\unitlength}{\svgwidth}%
  \fi%
  \global\let\svgwidth\undefined%
  \global\let\svgscale\undefined%
  \makeatother%
  \begin{picture}(1,0.41115656)%
    \lineheight{1}%
    \setlength\tabcolsep{0pt}%
    \put(0.69218917,0.01145174){\color[rgb]{0,0,0}\makebox(0,0)[lt]{\lineheight{1.25}\smash{\begin{tabular}[t]{l}(c) test set\end{tabular}}}}%
    \put(0.35660851,0.01076877){\color[rgb]{0,0,0}\makebox(0,0)[lt]{\lineheight{1.25}\smash{\begin{tabular}[t]{l}(b)  validation set\end{tabular}}}}%
    \put(0.02561126,0.01131267){\color[rgb]{0,0,0}\makebox(0,0)[lt]{\lineheight{1.25}\smash{\begin{tabular}[t]{l}(a) training set\end{tabular}}}}%
    \put(0,0){\includegraphics[width=\unitlength,page=1]{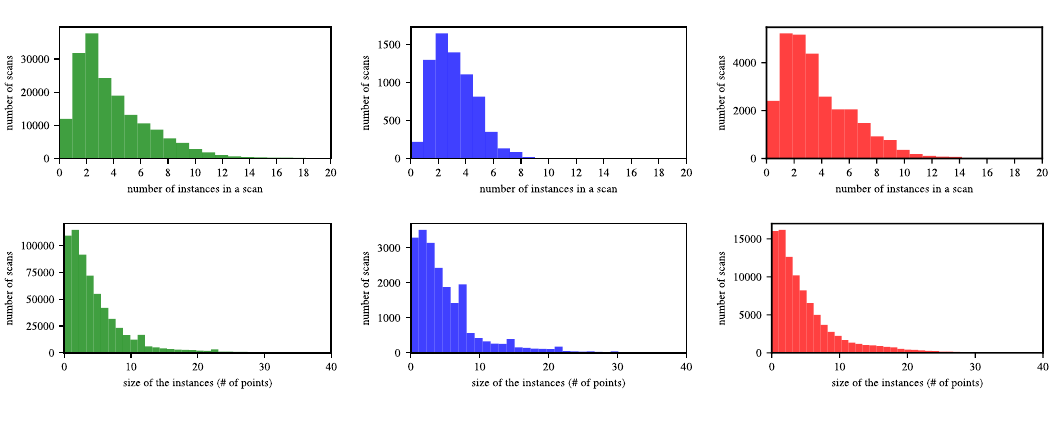}}%
  \end{picture}%
\endgroup%

 \caption{Statistics of the moving instance segmentation benchmark based on the RadarScenes dataset. The statistics are conducted individually for the training (a), the validation (b), and the test set (c). The top row illustrates the number of instances within a scan, and the bottom row shows the size of the instances, which correspond to the number of points. }
 \vspace{-0.1cm}
 \label{fig:stat}
  \vspace{-0.2cm}
\end{figure*}
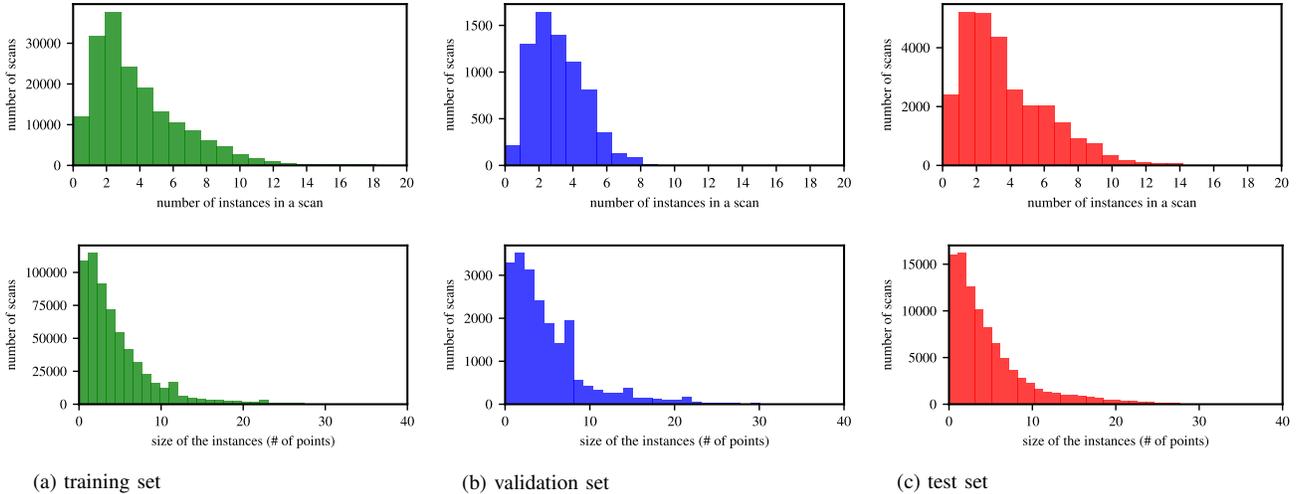
\subsection{Experimental Setup}
We perform our evaluation on RadarScenes~\cite{schumann2021icif}, the only large-scale open-source high-resolution radar dataset~\cite{zhou2022sensors} including precise point-wise annotations for moving detections for varying scenarios and different weather conditions. Additionally, we evaluate our method on the medium-sized View-of-Delft dataset~\cite{palffy2022ral} to illustrate the generalization capabilities of our method. For RadarScenes, we follow Zeller~\etalcite{zeller2023ral} for the training, validation, and test split of the 158 annotated sequences. We utilize the validation set to perform the ablation studies. Furthermore, we assemble the annotations to establish the new task of moving instance segmentation, which we introduce in~\secref{sec:bench}. 
For the View-of-Delft dataset, we keep the original training and validation split. We transfer the bounding box labels into point-wise annotations, including the differentiation between moving and static instances.
\begin{table}[t]
  \centering
{%
\begin{tabular}{@{}lccc@{}}
\toprule
RadarScenes~\cite{schumann2021icif}         & training set & validation set & test set \\ \midrule
number of scans            & $175,918$       & $7,062$           & $27,849$    \\
number of moving instances & $630,851$       & $21,143$          & $95,392$    \\
avg. number of points per scan & $569$          &       $559$           &   $534$          \\ \bottomrule
\end{tabular}%
}
\caption{The number of scans, the number of moving instances, and the average number of points per scan in the RadarScenes training, validation, and test set.}
  \label{tab:stat}
  \vspace{-0.2cm}
\end{table}
We utilize the panoptic quality~($\mathrm{PQ}$)~\cite{behley2021icra} to evaluate the combined task of semantic segmentation and instance segmentation. Additionally, we report the segmentation quality~($\mathrm{SQ}$), the recognition quality~($\mathrm{RQ}$), and the intersection over union~($\mathrm{IoU})$. To derive a detailed evaluation, we further differentiate between the static~(stat) and moving~(mov) classes.
\begin{table*}[t]
 \centering
\resizebox{\textwidth}{!}{%
\begin{tabular}{lcccc|cccc|cccc} 
\toprule
model                                & $\mathrm{PQ}$ & $\mathrm{mIoU}$ & $\mathrm{SQ}$ & $\mathrm{RQ}$ & $\mathrm{PQ}^{\mathrm{mov}}$ & $\mathrm{IoU}^{\mathrm{mov}}$ & $\mathrm{SQ}^{\mathrm{mov}}$ & $\mathrm{RQ}^{\mathrm{mov}}$ & $\mathrm{PQ}^{\mathrm{stat}}$ & $\mathrm{IoU}^{\mathrm{stat}}$ & $\mathrm{SQ}^{\mathrm{stat}}$ & $\mathrm{RQ}^{\mathrm{stat}}$  \\ 
\midrule
Threshold $\lvert v_i\rvert >v_t$ + HDBSCAN~\cite{campello2013dbscan}                                & 59.7 & 65.2 &       88.3         &64.6  &     23.7&   35.1   &     80.9 &  29.3  &   95.7  &   95.2    &  95.7   &    100.0               \\           
Threshold $\lvert v_i\rvert >v_t$ + mean shift~\cite{comaniciu2002tpami}                    & 57.2 & 65.2 &       90.4         &61.0  &     18.7&   35.1   &     85.1 &  22.0  &   95.7  &   95.2    &  95.7   &    100.0               \\
4DMOS~\cite{mersch2022ral} + HDBSCAN~\cite{campello2013dbscan}                       & 71.3          & 86.0          & 93.1          & 75.0          & 43.7              & 73.1               & 87.3              & 50.1              & 98.9               & 99.0                & 98.9               & 100.0               \\
4DMOS~\cite{mersch2022ral} + mean shift~\cite{comaniciu2002tpami}                    & 79.0 & 86.0 &       95.9         &81.9   &     59.1&   73.1   &     92.7 &  63.7  &   98.9  &   99.0    &  98.9   &    100.0               \\
Point Voxel Transformer~\cite{zhang2022ijis} + HDBSCAN~\cite{campello2013dbscan}    & 70.3          & 84.7          & 91.8          & 74.6          & 41.9              & 70.7               & 84.9              & 49.3              & 98.7               & 98.8                & 98.7               & 100.0               \\
Point Voxel Transformer~\cite{zhang2022ijis} + mean shift~\cite{comaniciu2002tpami} & 76.3          & 84.7         & 94.6          & 79.4          & 53.9             & 70.7               & 90.7              & 59.9              & 98.7               & 98.8                & 98.7               & 100.0            \\
Stratified Transformer~\cite{xin2022cvpr} + HDBSCAN~\cite{campello2013dbscan}     & 71.2          & 87.3          & 92.1          & 75.5          & 43.5              & 75.2              & 85.2              & 51.1              & 98.9               & 99.0                & 98.9               & 100.0               \\
Stratified Transformer~\cite{xin2022cvpr} + mean shift~\cite{comaniciu2002tpami}  & 78.4          & 87.3         & 94.8          & 81.7          & 57.9             & 75.2               & 90.8              & 63.7              & 98.9             &  99.0              & 98.9               & 100.0             \\
Gaussian Radar Transformer~\cite{zeller2023ral} + HDBSCAN~\cite{campello2013dbscan}     & 73.2          & 89.1         & 92.9          & 77.3      & 47.3              & 79.1               & 86.8              & 54.5             & 99.0             & 99.1                & 99.0               & 100.0               \\
Gaussian Radar Transformer~\cite{zeller2023ral} + mean shift~\cite{comaniciu2002tpami}  & 81.1          & 89.1         & 95.9          & 84.1       & 63.3              & 79.1               & 92.7              & 68.2              & 99.0             & 99.1                & 99.0               & 100.0              \\
DS-Net~\cite{hong2021cvpr}                               & 62.0          & 82.2          & 94.2          & 64.4          & 25.9              & 66.1               & 90.3              & 28.7              & 98.1               & 98.2                & 98.1               & 100.0               \\ 
Mask3D~\cite{schult2022arxiv}                               & 80.4          & 84.3          & 96.4          & 83.0          & 62.0             & 69.8               & 94.0              & 66.0              & 98.8               & 98.9                & 98.8               & 100.0               \\ 

\midrule
Ours                                 & \textbf{86.5}          & \textbf{92.6}       & \textbf{96.9}          & \textbf{89.0}      & \textbf{73.6}              & \textbf{85.7}             & \textbf{94.4}            & \textbf{78.0}             & \textbf{99.4}            & \textbf{99.4 }             & \textbf{99.4 }            & \textbf{100.0}               \\
\bottomrule
\end{tabular}
}
\caption{Moving instance segmentation results on the RadarScenes test set in terms of $\mathrm{PQ}$, $\mathrm{SQ}$, $\mathrm{RQ}$, and $\mathrm{IoU}$ scores. 
  \vspace{-0.2cm}}
  \label{tab:resall}
\end{table*}
\subsection{Moving Instance Segmentation Benchmark}
\label{sec:bench}
The RadarScenes~\cite{schumann2021icif} dataset consists of 158 annotated sequences split into 130 sequences for training and 28 sequences for testing. We utilize six of the 28 sequences as a validation set (sequences: 6, 42, 58, 85, 99, 122) as previously done~\cite{zeller2023ral}. 

The RadarScenes dataset provides individual point clouds for the four radar sensors. The measurements are from the near-range mode of the $\SI{77}{GHz}$ automotive radar sensors, which cover detections in a range of up to $\SI{100}{m}$. Two sensors are mounted at $\pm \SI{85}{\degree}$ and two sensors at $\pm \SI{25}{\degree}$ with respect to the driving direction. Since the directions in which the sensors point are different, we combine the individual point clouds into one scan based on the relative transformations to obtain information about the surroundings of the vehicle following Zeller~\etalcite{zeller2023ral}. However, at the beginning of the sequence, we recognized that the recordings were asynchronous. To account for that, we only combine the intermediate point clouds of the individual sensors until the next measurement of an already included sensor is available. The reason is that if we include multiple point clouds of one sensor, the pose extraction is more erroneous because each point cloud has corresponding pose information. Since the resulting point clouds are recorded within a short time frame, we do not compensate for additional movement.

We extract the time and the pose information based on the first measurement included within the combined scan in the common coordinate system. The scenes are represented as a sequence of scans where each scan includes up to four radar sensors. The semantic annotations for the moving agents comprise the car, large vehicle, two-wheeler, pedestrian, and pedestrian group class. We transfer the track IDs into instance labels for the individual scans. 

We summarize the number of scans for the individual dataset splits in~\tabref{tab:stat}. With more than $200,000$ scans in total, the large-scale RadarScenes dataset enables data-driven moving instance segmentation. To illustrate the diversity of the dataset, we show in~\figref{fig:stat} histograms of the number of instances within a scan and the size of the instances, which is equal to the number of points per instance. The number of instances within a scan displays the diversity of scenes with changing numbers of agents. Furthermore, the small size of the instances illustrates the difficulty of reliably detecting moving agents within sparse radar point clouds. Moreover, 87\% of the points with an absolute ego-motion compensated Doppler velocity larger than \SI{0.1}{m/s} are noise and hence belong to the static class. The sparsity and noisiness demonstrate that reliable moving instance segmentation in radar data is more challenging compared to typically used 3D LiDAR data. To enable further research, the benchmark will be made available upon acceptance.

\begin{figure}[t]
  \centering
  \fontsize{8pt}{8pt}\selectfont
     \def\svgwidth{\linewidth}
     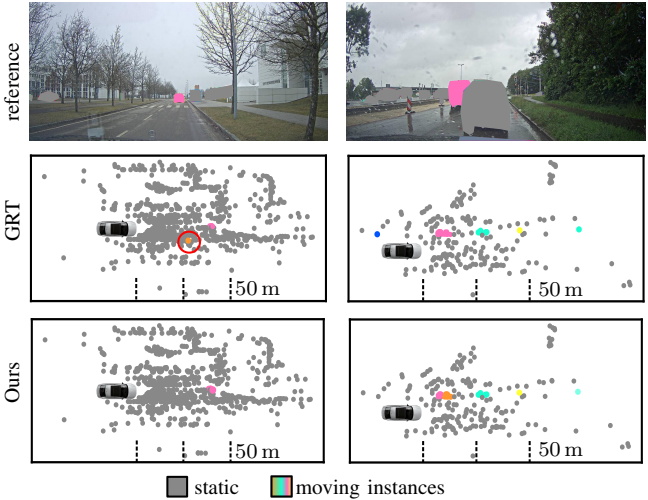
  \caption{Qualitative results of our Radar Instance Transformer and the Gaussian Radar Transformer~(GRT)~\cite{zeller2023ral} on the test set of RadarScenes. The camera images are anonymized and just shown for reference. The colors in the image do not correspond to the predicted instances. The left column is from sequence 14 (fog), and the right is from sequence 93 (rain). In the images of the predictions, each color represents a different instance of moving objects (static is grey). The colors in the images correspond if the object is visible.}
  \vspace{-0.2cm}
  \label{fig:qres}
  \vspace{-0.2cm}
\end{figure}
\begin{figure}[t]
  \centering
  \fontsize{8pt}{8pt}\selectfont
     \def\svgwidth{\linewidth}
     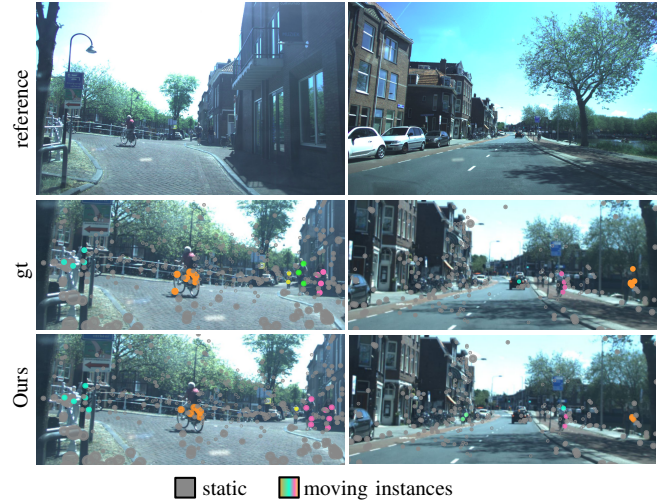
  \caption{Qualitative results of our Radar Instance Transformer on the validation set of View-of-Delft. The camera images are shown for reference. The ground truth (gt) instance labels and the corresponding instance predictions are projected into the reference image. The ground truth and the predictions are cropped for better visibility.}
  \vspace{-0.2cm}
  \label{fig:qresvod}
  \vspace{-0.2cm}
\end{figure}

\subsection{Moving Instance Segmentation}
\label{sec:mis}
The first experiment evaluates the performance of our approach, and its outcome supports the claim that we achieve state-of-the-art results for moving instance segmentation in sparse and noisy radar point clouds.
We compare our Radar Instance Transformer to the recent and high-performing networks with a strong performance on different benchmarks, including moving object~\cite{mersch2022ral}, semantic~\cite{xin2022cvpr,zeller2023ral,zhang2022ijis}, instance~\cite{schult2022arxiv}, and panoptic segmentation~\cite{hong2021cvpr}. We added the two approaches~\cite{hong2021cvpr,schult2022arxiv} based on the fact that the clustering has to be class-agnostic because all instances belong to the moving class. Additionally, the methods do not include range image representations since RadarScenes does only provide x and y coordinates. To derive panoptic labels, we extend Mask3D~\cite{schult2022arxiv} with state-of-the-art post-processing for mask predictions~\cite{cheng2022cvpr}. We extend the semantic and moving object segmentation approaches~\cite{mersch2022ral,xin2022cvpr,zeller2023ral,zhang2022ijis} with commonly used clustering algorithms, namely HDBSCAN~\cite{campello2013dbscan} and mean shift~\cite{comaniciu2002tpami}, to group points into instances. Furthermore, we add a threshold-based baseline that first identifies moving detections based on the ego-motion compensated Doppler velocities where threshold \mbox{$v_t=0.92\,m/s$} and utilizes the clustering algorithms to group the detected point into moving instances. To reliably detect the most vulnerable road users, the pedestrians, we have to identify instances that comprise just a single detection. Therefore, we optimize the hyperparameters and set the minimum cluster size for HDBSCAN to 1. For mean shift, we optimize the bandwidth~$b^{ms}$ based on the overall performance on the validation set, resulting in 3.5, 4.5, 5.0, and 4.5 for 4DMOS~\cite{mersch2022ral}, Stratified Transformer~\cite{xin2022cvpr}, Point Voxel Transformer~\cite{zhang2022ijis} and Gaussian Radar Transformer~\cite{zeller2023ral}, respectively. To reliably detect larger objects, we additionally implement an offset prediction head~\cite{hong2022arxiv} to support the clustering and enhance accuracy for point-based methods. The offset prediction head~\cite{hong2022arxiv} combines two fully connected layers, batch normalization~\cite{ioffe2015arxiv}, and a rectified linear unit~(ReLU)~\cite{nair2010icml} to regress offset to the instance centers. We concatenate the features of the respective backbone $\mathbf{X}^b$ and the coordinates $\mathbf{P}^b$ to include fine-grained position information. The offsets $\mathbf{O} \in \RR^{N\times 3}$ point from the point coordinates $\mathbf{P} \in \RR^{N \times 3}$ to the instance centers $C\in \RR^{N \times 3}$. We derive the center based on the ground truth instance IDs and point coordinates. The loss function for offset regression can be expressed as follows:
\begin{equation}
    L^{\mathrm{off}} = \frac{1}{N}\sum_{i=1}^{N}\lVert \mathbf{o}_i - (\mathbf{c}_i - \mathbf{p}_i) \rVert_1\text{,}
\end{equation}
where $N$ is the number of points in the current point cloud. 

Our Radar Instance Transformer outperforms the existing methods, especially in terms of $\mathrm{PQ}$ and $\mathrm{IoU}$, as displayed in \tabref{tab:resall}. The threshold-based baseline further illustrates the difficulties of the task and underlines the necessity of advanced approaches. The learning-based approaches improve $\mathrm{IoU}^{\mathrm{mov}}$ by more than 30 absolute percentage points. DS-Net utilizes an optimized instance clustering, but point-based approaches~\cite{xin2022cvpr,zhang2022ijis} clearly exceed the voxel-based method, which underlines that minimizing discretization artifacts enhances performance. Point Voxel Transformer includes a point-based branch, which helps to reduce the negative effect. \begin{table}[t]
  \centering
  \setlength\tabcolsep{3pt}
{%
\begin{tabular}{@{}lllll@{}}
\toprule
model                       & $\mathrm{PQ}^{\mathrm{mov}}$            & $\mathrm{SQ} ^{\mathrm{mov}}$            & $\mathrm{RQ}^{\mathrm{mov}}$             & $\mathrm{IoU}^{\mathrm{mov}}$             \\ \midrule
Stratified Transformer~\cite{xin2022cvpr}     & 22.8   & 78.1 & 29.3 & 55.6 \\
Gaussian Radar Transformer~\cite{zeller2023ral} &  26.0  & \textbf{79.4}  & 32.7 & 62.3\\
Ours                       & \textbf{26.9} & 79.1 & \textbf{34.0} & \textbf{63.3}       \\ \bottomrule
\end{tabular}%
}
\caption{Moving instance segmentation results on the View-of-Delft~\cite{palffy2022ral} validation set in terms of $\mathrm{PQ}^{\mathrm{mov}}$, $\mathrm{SQ} ^{\mathrm{mov}}$, $\mathrm{RQ}^{\mathrm{mov}}$, and $\mathrm{IoU}^{\mathrm{mov}}$ scores.}
  \label{tab:vod}
  \vspace{-0.2cm}
\end{table}Contrary, 4DMOS~\cite{mersch2022ral} enhances performance by aggregating scans and prediction smoothing over time. However, the instance assignment is difficult. We assume that the problems result from the fact that all instances belong to the same class, which makes offset prediction and clustering more complex. Nevertheless, algorithms that do not depend on semantic labels are more generally applicable and potentially cover the long-tail classes. The dedicated mask predictions of Mask3D~\cite{schult2022arxiv} overcome this limitation and improve $\mathrm{PQ}$. However, our Radar Instance Transformer enhances $\mathrm{PQ}^{\mathrm{mov}}$ by 10 absolute percentage points compared to Mask3D, which is a strong improvement.
The Stratified Transformer performs well without temporal information. Particularly the mean shift algorithm performs well since the restriction of a cluster of one for HDBSCAN makes it difficult to cluster larger objects reliably. However, even in the semantic domain, our Radar Instance Transformer enhances $\mathrm{IoU}^{\mathrm{mov}}$ by more than 9 absolute percentage points, which again illustrates the superiority of our method. Moreover, we even surpass the radar-specific method, the Gaussian Radar Transformer~\cite{zeller2023ral}, by more than 5 absolute percentage points in terms of $\mathrm{IoU}^{\mathrm{mov}}$.

To further illustrate the generalization capability of our model, we evaluate the three best-performing methods in terms of $\mathrm{IoU}^{\mathrm{mov}}$ on the View-of-Delft validation set. \tabref{tab:vod} shows the superior performance of our method compared to the Gaussian Radar Transformer and Stratified Transformer. To enable a fair comparison, we utilize the $\mathrm{IoU}$ since both baselines are developed for semantic segmentation. The Stratified Transformer achieves an $\mathrm{IoU}^{\mathrm{stat}}$ of $98.1$, whereas the Gaussian Radar Transformer and our method achieve an $\mathrm{IoU}^{\mathrm{stat}}$ of $98.4$. Nevertheless, our approach enhances the $\mathrm{IoU}^{\mathrm{mov}}$ by 1 absolute percentage point compared to the best-performing radar-specific method and by more than 7 absolute percentage compared to Stratified Transformer. We observe that the relative transformations to align the point clouds are more precise in the RadarScenes dataset. We assume that this might be one reason why the improvements are more limited, as detailed in~\secref{sec:absafe}. Overall, the approaches perform well in both moving instance segmentation benchmarks, which is a good starting point for future research. One possibility is to predict the semantic labels in a cluster-then-classify approach, which was out of the scope of this paper.

\begin{figure*}[ht]
 \centering
 \fontsize{8pt}{8pt}\selectfont
 \def\svgwidth{0.94\textwidth}
 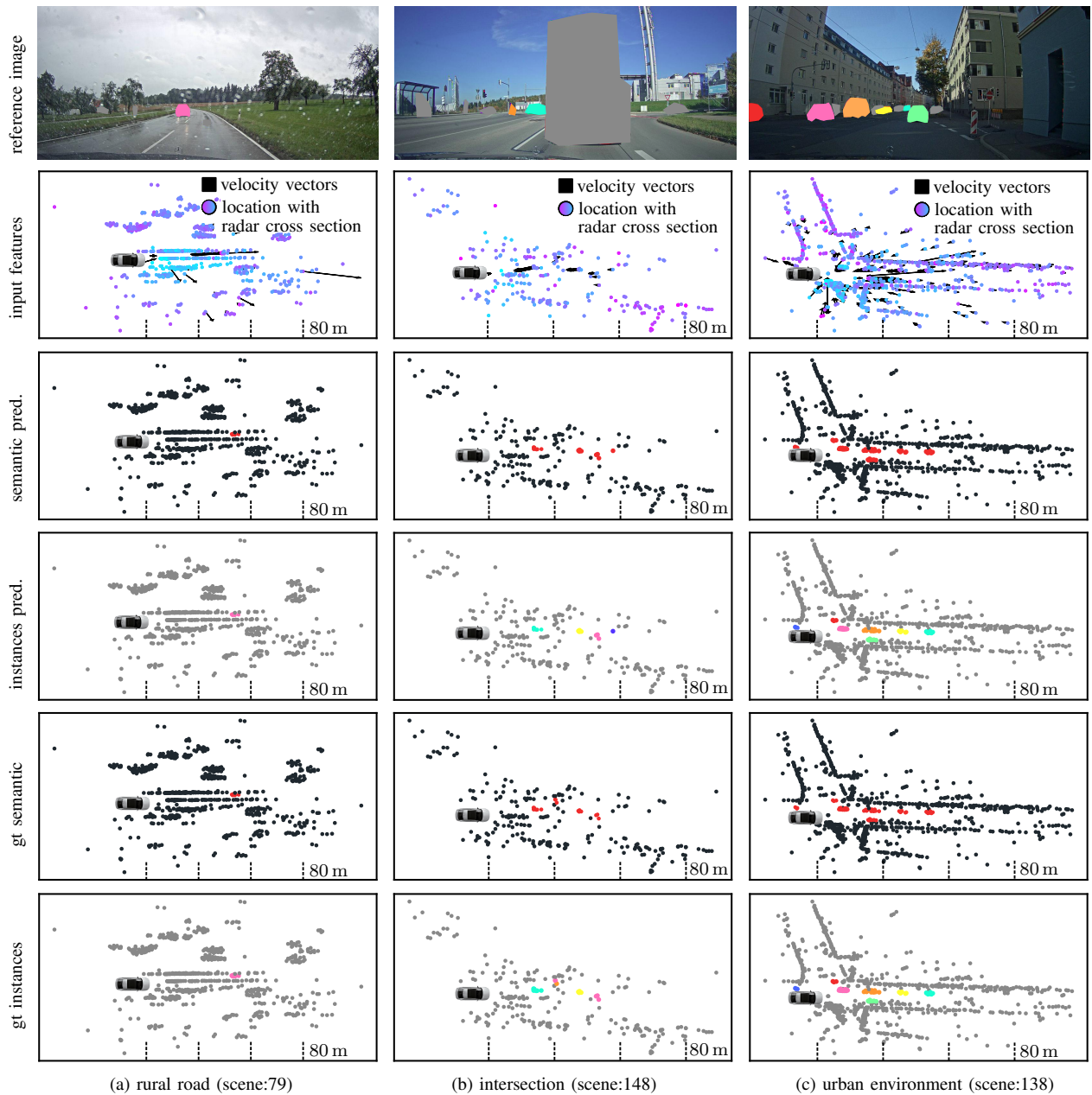
 \caption{Qualitative results of our Radar Instance Transformer on the RadarScenes test set. The reference camera images are anonymized. The colors in the illustration of the instance predictions~(pred.) and the ground truth~(gt) instances have the same color. The colors in the image correspond if the object is visible. The results illustrate the performance within different scenes, including a rural road on a rainy day (a), an intersection with agents crossing (b), and an urban environment with multiple agents (c).}
 \label{fig:arappres}
\end{figure*}

\figref{fig:qres} shows some qualitative results on the RadarScenes test set of our approach and the Gaussian Radar Transformer. Our approach correctly identifies the distant moving instance in scene 14 and does not include a false positive prediction. Furthermore, our Radar Instance Transformer works reliably under versatile scenes, as illustrated in~\figref{fig:arappres}. Our approach reliably identifies moving instances under different and changing scenarios. However, for the intersection scene, the measured Doppler velocities of the agents are small, and the reliable identification of all distant moving instances is difficult. Since the Doppler velocity is the radial velocity of the detections, the tangential movement is not covered. Despite that fact, the network accurately differentiates between moving and static agents. Furthermore, our approach identifies the oncoming car in rainy scene 79 precisely. The scene illustrates the handling of noise, which is best seen for the Doppler velocity vectors for off-road detections belonging to the static class, as depicted in the ground truth. Additionally, our approach identifies multiple moving agents in the urban environment. The different scenarios further illustrate the sparsity and the changing density of the radar scan. \figref{fig:qresvod} illustrates further qualitative results on the View-of-Delft validation set of our approach. Overall, our approach performs well despite the various difficulties.
\begin{table}[t]
  \centering
{%
\begin{tabular}{@{}lccccc@{}}
\toprule
\multicolumn{1}{l}{} & PC  & $\mathrm{PQ}^{\mathrm{mov}}$            & $\mathrm{SQ} ^{\mathrm{mov}}$            & $\mathrm{RQ}^{\mathrm{mov}}$             & $\mathrm{IoU}^{\mathrm{mov}}$             \\ 
\midrule
$T=0$  &      \checkmark          & 73.8         & 94.7          & 77.9          & 82.2           \\
$T=1$   &      \checkmark        & 74.0          & 94.8          & 78.0         & 82.5           \\
$T=3$    &     \checkmark          & 76.4          & 95.0         & 80.4          & 83.6           \\ 
$T=4$   &      \checkmark        & 77.3          & 95.2         & 81.2         & \textbf{84.8}           \\
$T=5$    &     \checkmark          & \textbf{77.7}          & 95.2          & \textbf{81.6}         & 84.5           \\ 
\midrule

Ours without PC &    & 76.1          & \textbf{95.3}          & 79.8          & 83.3           \\ 
\midrule
Ours ($T=2$)  &   \checkmark     & 76.8 & 95.1 & 80.8 & 84.4  \\
\bottomrule
\end{tabular}
}
\caption{Influence of the number of previous input point clouds and pose compensation~(PC) on the RadarScenes validation set.}
  \label{tab:safe}
  \vspace{-0.1cm}
\end{table}
\subsection{Ablation Studies on the SAFE Module}
\label{sec:absafe}
The second experiment evaluates our sequential attentive feature encoding and illustrates that our module improves performance by efficiently including valuable temporal information within the features of the current point cloud. 
For this experiment, we vary the number of input scans for our Radar Instance Transformer and evaluate the pose compensation as depicted in~\tabref{tab:safe}. The temporal information improves the performance with the best result for five additional previous scans, \ie, $T=5$. However, the training time, memory consumption, and runtime increase with adding more additional scans. Therefore, we select $T=2$ as the best trade-off since our method performs well in terms of $\mathrm{PQ}^{\mathrm{mov}}$ and $\mathrm{IoU}^{\mathrm{mov}}$ and reduces the inference time. Furthermore, to save resources, the training time decreases by more than a factor of two compared to $T=4$.
The enhancement by temporally enriched features is directly visible by the increase of more than 2 absolute percentage points by including $T=2$ previous scans instead of none. Furthermore, we argue that recent scans include the most important information to identify moving instances, leading to a slight decrease in accuracy for $T=3$. Additionally, the pose compensation does not compensate for the relative movement of instances, leading to false associations over time. Thus, the recent scans contain the most valuable information, and the $\mathrm{IoU}^{\mathrm{mov}}$ shows just small improvements. To verify that the temporal information supports the identification of static points, including noise, we determine the $\mathrm{IoU}^{\mathrm{stat}}$ for $T=0$ and $T=2$ resulting in 99.4 and 99.5, respectively. We assume that the changing appearance of noise can be identified within the temporal domain, and hence, $\mathrm{IoU}^{\mathrm{stat}}$ improves. To elaborate on the improvement in detail, separate labels for static and noise are required, which are not available.

In the second step, we remove the pose compensation, which helps to align the previous and the current scan. Here, the decrease in $\mathrm{PQ}^{\mathrm{mov}}$ is small. One advantage of radar sensors is the often higher frame rate compared to LiDAR sensors, which leads to smaller transformations between scans. In addition, we expect that the self-attention mechanism adaptively controls the information exchange and helps to extract valuable features, which underlines the advantages of our SAFE module.

\begin{table}[t]
  \centering
  \begin{tabular}{@{}lllcc@{}}
\toprule
\#    & \begin{tabular}[c]{@{}l@{}}local similarity \\($\mathbf{S}^{\mathrm{loc}}$)\end{tabular} & \begin{tabular}[c]{@{}l@{}}global similarity\\($\mathbf{S}^{\mathrm{glob}}$)\end{tabular} & $\mathrm{IoU}^{\mathrm{mov}}$            & $\mathrm{PQ}^{\mathrm{mov}}$              \\ 
\cmidrule{1-5}
{[}A] &                                                             &                                                             & 81.6          & -              \\
{[}B] & \multicolumn{1}{c}{\checkmark }                                                            &                                                             & 83.7          & -              \\
{[}C] &                                                             & \multicolumn{1}{c}{\checkmark }                                                            & 83.2          & 75.6           \\
{[}D] &\multicolumn{1}{c}{\checkmark }                                                            & \multicolumn{1}{c}{\checkmark }                                                            & \textbf{84.4} & \textbf{76.8}  \\ 
\midrule
\multicolumn{3}{l}{global: with positional encoding }                                                                                   & 82.6          & 74.2          \\
\multicolumn{3}{l}{local: static included}                                                                                               & 82.7          & 74.0          \\
\bottomrule
\end{tabular}
\caption{Influence of the different components of our instance transformer head in terms of $\mathrm{IoU}^{\mathrm{mov}}$ and $\mathrm{PQ}^{\mathrm{mov}}$ on the RadarScenes validation set.}
  \label{tab:componentsith}
  \vspace{-0.2cm}
\end{table}
\subsection{Ablation Studies on the Instance Transformer Head}
\label{sec:abltranshead}
The third experiment is presented to illustrate that our instance transformer head incorporates important instance information and enhances the performance of moving instance segmentation. Notably, we do not report the $\mathrm{PQ}^{\mathrm{mov}}$ for the models without the global attentive similarity quantity because we cannot perform the partitioning based on the learned weights, and thus we compare the results for this experiment based on the $\mathrm{IoU}^{\mathrm{mov}}$. 
To assess the benefits of the proposed attentive similarity quantities, we remove the respective modules for the global and local computation, as depicted in~\tabref{tab:componentsith}.
We first remove the local and global attentive similarity in configuration [A], which reduces the $\mathrm{IoU}^{\mathrm{mov}}$ by 2.8 absolute percentage points. In ablation [B], we add the local attention, which already improves the $\mathrm{IoU}^{\mathrm{mov}}$ compared to [A]. We assume that the knowledge of which point within the local neighborhood belongs to the same instance helps to differentiate between instances and also to identify moving detections reliably. Compared to [A], the added global attentive similarity in [C] improves the performance. The global context is beneficial for identifying moving instances. Furthermore, false positive detections of moving detection are penalized, which enhances moving object segmentation. The combined local and global attentive similarity [D] combines both properties and further improves $\mathrm{PQ}^{\mathrm{mov}}$. 

To analyze our design decisions, we add the positional encoding for the global attentive similarity. Since the relative position grows large in magnitude in the global context, as mentioned in~\secref{sec:mith}, we assume that the local context is lost, which is essential to differentiate between nearby points and hence does not improve accuracy. Additionally, we exclude the static point from the local similarity in our method, which means that for all static points, the predicted attentive similarity quantity has to be 0. We presume that the focus is clearly on moving instances, which is not the case if we include the static points and thus decrease $\mathrm{PQ}^{\mathrm{mov}}$. In conclusion, the instance transformer head improves performance by exploiting local and global context to extract valuable features from noisy radar point clouds.

\begin{table}[t]
  \centering
{%
\begin{tabular}{@{}lccc@{}}
\toprule
\multicolumn{1}{l}{} & $\mathrm{PQ}^{\mathrm{mov}}$             & $\mathrm{SQ}^{\mathrm{mov}}$             & $\mathrm{RQ}^{\mathrm{mov}}$              \\ 
\midrule

r=3.0                & 70.4          & 94.5          & 74.5           \\
r=4.0                & 73.4          & 95.3          & 77.0           \\
r=5.0                & 75.1          & \textbf{95.4}  & 78.7           \\
r=6.5                & 76.6          & 95.1         & 80.6           \\
r=7.5                & \textbf{76.8} & 95.1 & \textbf{80.8} \\ 
r=8.0                & 76.7          & 95.0         & \textbf{80.8}         \\ 
\midrule
Ours (r=7.0)         & \textbf{76.8} & 95.1 & \textbf{80.8}  \\ \bottomrule
\end{tabular}}
  \caption{Ablation study on the graph-based instance assignment on the RadarScenes validation set. \vspace{-0.1cm}}
  \label{tab:abgraph}
\end{table}
\subsection{Ablation Studies on the Instance Assignment}
The ablation study presented in this section supports our claim that attention-based graph partitioning enhances instance assignments without class-dependent information. We evaluate the hyperparameter $r$ for the radius graph, as depicted in~\tabref{tab:abgraph}. Our method achieves good performance for different radii with the best $\mathrm{PQ}^{\mathrm{mov}}$ for $r=\SI{7}{m}$. Compared with the bandwidths for mean shift algorithms in~\secref{sec:mis}, our approach is able to utilize a larger radius. One reason for a larger radius is that we do not use an offset prediction and still want to reliably detect large moving instances. But this could lead to a combination of instances, especially small instances with nearby larger instances. However, the major advantage of our attention-based graph partitioning is that the global attentive similarity quantity can effectively solve the interconnection between these instances. We additionally evaluated the performance of the most vulnerable road users, pedestrians, which often comprise single detections to show that the interconnection can be reliably solved. However, potential future work might address the changing resolution of radar point clouds based on the distance to the sensor to refine the instance assignment. Within the evaluated range of radii, the $\mathrm{PQ}^{\mathrm{mov}}$ for pedestrians remains stable at 84.7, which illustrates the strong capability of our approach. In comparison, clustering often relies on accurate offset predictions and class-specific properties~\cite{palffy2020ral} to reliably identify instances of varying size. Hence, our attention-based graph partitioning helps to identify moving instances reliably. 
\begin{table}[t]
  \centering
{%
\begin{tabular}{@{}lcc@{}}
\toprule
model                          & $\mathrm{IoU}^{\mathrm{mov}}$   & $\mathrm{PQ}^{\mathrm{mov}}$                     \\ \midrule
Ours + Point Voxel Transformer~\cite{zhang2022ijis} &   80.0   &  69.0                     \\
Ours + Stratified Transformer~\cite{xin2022cvpr}   & 82.2 & 70.1                  \\

Ours w/o top level             & 81.9 & 73.7                      \\ \midrule
Ours                           & \textbf{84.4} & \textbf{76.8}                  \\ \bottomrule
\end{tabular}%
}
\caption{Ablation study on the backbone design and comparison to recently best-performing backbones on the RadarScenes validation set.}
  \label{tab:backabl}
  \vspace{-0.2cm}
\end{table}
\subsection{Ablation Studies on the Backbone Architecture}
Finally, we analyze our method with respect to the ability to extract valuable features by keeping the full resolution of the point cloud throughout the network. The results in~\tabref{tab:backabl} verify that removing the intermediate transformer blocks on the top level of our model harms the accuracy. These building blocks are the additional modules compared to the widely-used U-Net~\cite{qi2017nips2,zhao2021iccv} architecture. Furthermore, we extend the two best-performing non-radar-specific backbones, which follow the U-Net architecture, with our model-agnostic modules to illustrate the advancements of our approach. The full-resolution processing and the extraction of additional high-level features outperform other methods. 
Nevertheless, our model-agnostic sequential attentive feature encoding and instance transformer head enhance the performance of the Stratified Transformer~\cite{xin2022cvpr} and the Point Voxel Transformer~\cite{zhang2022ijis}, which illustrates that these modules help to improve scene interpretation in radar scans.
\begin{table}[t]
  \centering
{%
\begin{tabular}{@{}lrr@{}}
\toprule
model                     & parameters (M) & mean runtime (ms) \\ \midrule
4DMOS~\cite{mersch2022ral} & 1.8           & 14.0         \\
Point Voxel Transformer~\cite{zhang2022ijis}   & 6.2            & 47.1        \\
Stratified Transformer~\cite{xin2022cvpr}    & 8.0            & 34.4        \\
DS-Net~\cite{hong2021cvpr}                    & 50.2           & 52.3         \\
Mask3d~\cite{schult2022arxiv}                    & 39.6           & 85.2         \\
Gaussian Radar Transformer~\cite{zeller2023ral}&       8.4         & 24.0         \\
\midrule
Ours                      & 3.8            & 31.7         \\
 \bottomrule
\end{tabular}%
}
\caption{The evaluation of the number of parameters and the mean runtime of the models on an Nvidia RTX A6000 GPU based on 1000 randomly sampled point clouds of the RadarScenes validation set.}
  \label{tab:runtimetabl}
  \vspace{-0.1cm}
\end{table}
\subsection{Runtime}
Finally, we analyze the runtime and the number of parameters of the different methods. We measured the runtimes on an AMD Threadripper CPU with an Nvidia RTX A6000 GPU.
We utilize an optimized farthest point sampling and $k$NN algorithm in C++ to speed up the inference of the algorithms. Since the point clouds differ in the number of detections, we evaluate 1,000 scans that are randomly selected from the validation set. For our approach, we utilize the best-performing configuration and include the $T=2$ previous scans. 

All results are detailed in~\tabref{tab:runtimetabl}.
The mean runtime of our approach is 31.7\,ms, which is equal to 31\,Hz, and thus faster than the frame rate of 17\,Hz of the sensor. The Gaussian Radar Transformer has the lowest runtime of the transformer-based models in our experiments with 24\,ms but utilizes twice as many parameters. We argue that the runtime depends on the different specific implementations of the respective attention mechanism. The Stratified Transformer~\cite{xin2022cvpr} uses a more complex mechanism that induces further latencies compared to the simple vector attention of the Gaussian Radar Transformer. Additionally, we observe that the runtime scales with the number of points. To illustrate that aspect, we calculated the minimum and maximum runtime of our model, which are 23\,ms and 221\,ms, respectively. For all models, scans without any moving object prediction do not need clustering, which reduces runtime. However, with the increasing number of moving predictions, our global attentive association matrix gets more complex, and therefore, the runtime scales with the number of points and moving predictions. Furthermore, the runtime scales with the number of $T$ previous scans, which increases to 35.3\,ms for $T=5$. Future research can address these issues to further decrease the runtime. Another option to reduce the runtime is the replacement of the commonly used kNN algorithm. Since the resolution of radar point clouds is dependent on the distance to the sensor, the adjustment of the sampling might address this limitation and additionally improve performance. 

To get an in-depth analysis, we additionally report the runtime and parameters of the individual blocks in~\tabref{tab:runtimetabl2}. We observe that the processing of the intermediate building blocks of our backbone is suboptimal, and additional explicit parallelization can enhance the inference time. Nevertheless, we achieve state-of-the-art performance with an adequate runtime, outperforming all other methods except Gaussian Radar Transformer and \mbox{DS-Net}, while we include temporal information and use fewer parameters. 

In summary, our evaluation suggests that our method
provides competitive moving instance segmentation in sparse radar
point clouds by efficiently including temporal information. The attentive similarity quantities encode valuable information and enable class-agnostic, graph-based instance assignment, outperforming state-of-the-art approaches. Thus, we support all our claims through this experimental evaluation.

\begin{table}[t]
  \centering
{%
\begin{tabular}{@{}lrr@{}}
\toprule
Stages of our model                    & parameters (M) & mean runtime (ms) \\ \midrule
Pre-processing            &  -             & 5.8          \\
SAFE                      &   0.006              & 2.7          \\
Backbone                  &  3.8              & 15.7         \\
Instance Transformer Head &   0.02            & 1.2          \\
Graph Partitioning          &    -            & 6.3          \\
 \bottomrule
\end{tabular}%
}
\caption{The detailed evaluation of the number of parameters and the mean runtime of the individual building blocks of our model on an Nvidia RTX A6000 GPU based on 1000 randomly sampled point clouds of the RadarScenes validation set.}
  \label{tab:runtimetabl2}
  \vspace{-0.2cm}
\end{table}
\section{Conclusion}
\label{sec:conclusion}

In this paper, we presented a novel approach for moving instance segmentation in sparse and noisy radar point clouds that shows strong performance and outperforms a large set of baseline methods. Our method efficiently exploits temporal information and overcomes the limitations of passing aggregated scans through the whole network. We utilize the self-attention mechanism throughout the network to extract valuable features and introduce attentive graph-based instance partitioning. This allows us to successfully identify moving agents and enhance the feature extraction by incorporating local and global instance knowledge. We furthermore established a new benchmark based on the RadarScenes dataset, which allows further comparisons with future work, providing comparisons to other methods and supporting all claims made in this paper. 
The experiments suggest that the different parts of our approach are essential to achieve good performance on moving instance segmentation.
Overall, our approach outperforms the state of the art, taking a step forward towards reliable moving instance segmentation and sensor redundancy for autonomous vehicles and robots.

\small
\bibliographystyle{plain_abbrv}
\bibliography{ref_rss}

\end{document}